\documentclass[runningheads]{llncs}
\usepackage[T1]{fontenc}
\usepackage{graphicx}
\usepackage{amsmath}
\usepackage{amssymb}
\usepackage{hyperref} 
\begin{document}
\title{APTx Neuron: A Unified Trainable Neuron Architecture Integrating Activation and Computation}

\titlerunning{APTx Neuron}

\author{Ravin Kumar \orcidID{0000-0002-3416-2679}}
\authorrunning{Ravin Kumar}
%
\institute{Department of  Computer Science, Meerut Institute of Engineering and Technology, Meerut-250005, Uttar Pradesh, India \\
\email{ravin.kumar.cs.2013@miet.ac.in}}
\maketitle              
\begin{abstract}
We propose the APTx Neuron, a novel, unified neural computation unit that integrates non-linear activation and linear transformation into a single trainable expression. The APTx Neuron is derived from the APTx activation function, thereby eliminating the need for separate activation layers and making the architecture both optimization-efficient and elegant. The proposed neuron follows the functional form $y = \sum_{i=1}^{n} ((\alpha_i + \tanh(\beta_i x_i)) \cdot \gamma_i x_i) + \delta$, where all parameters $\alpha_i$, $\beta_i$, $\gamma_i$, and $\delta$ are trainable. We validate our APTx Neuron-based architecture on the MNIST dataset, achieving up to $96.69\%$ test accuracy within 11 epochs using approximately 332K trainable parameters. The results highlight the superior expressiveness and training efficiency of the APTx Neuron compared to traditional neurons, pointing toward a new paradigm in unified neuron design and the architectures built upon it. Source code is available at \url{https://github.com/mr-ravin/aptx_neuron}.
\keywords{APTx Neuron \and APTx Activation Function \and Activation Function \and Unified Neuron \and Deep Learning \and MNIST}
\end{abstract}
\section{Introduction}
Custom activation functions such as Swish~\cite{ramachandran2017searchingactivationfunctions}, Mish~\cite{misra2019mish}, and ELU~\cite{clevert2015fast} have demonstrated superior performance compared to the traditional ReLU~\cite{10.5555/3104322.3104425} in various deep learning applications. These advancements reflect a broader trend toward more adaptive and expressive non-linearities. 

Among recent innovations, the APTx activation function~\cite{kumar2022aptx,aptxactivation2022code} is notable for its parametric, trainable formulation, which can approximate multiple activation behaviors, including Swish and Mish, and can also resemble Tanh-like curves under certain parameter settings. As noted in recent surveys on trainable activation functions~\cite{APICELLA202114}, such flexibility enables neural networks to better adapt to task-specific requirements during training. The mathematical formulation of the APTx activation function is given in Equation~\ref{eq:aptx_activation}.

\begin{equation}\label{eq:aptx_activation}
y = (\alpha + \tanh(\beta x)) \cdot \gamma x
\end{equation}
In Equation~\ref{eq:aptx_activation}, $\alpha$, $\beta$, and $\gamma$ are learnable parameters. This formulation enables adaptation of the dynamic shape during training for the APTx activation function. 

Notably, the APTx activation function subsumes several well-known activations as special cases or close approximations. For example, it exactly recovers the SWISH($x,\rho$) function when $\alpha = 1$, $\beta = \rho/2$, and $\gamma = 1/2$. Similarly, it can closely approximate the MISH function by setting $\alpha = 1$, $\beta = 1/2$, and $\gamma = 1/2$ for the negative domain, and $\alpha = 1$, $\beta = 1$, and $\gamma = 1/2$ for the positive domain.

The ReLU activation function can be approximated using the APTx activation function by fixing $\alpha = 1$ and $\gamma = 1/2$, with the approximation improving as $\beta$ increases and converging to ReLU in the limit $\beta \to \infty$. In practice, setting $\alpha = 1$, $\beta \approx 10^{6}$, and $\gamma = 1/2$ already produces a close approximation of ReLU.

In this work, we extend the APTx function from just an activation function to a full-fledged trainable neuron by adding a bias term $\delta$ and integrating the summation mechanism of a neuron. The result is a compact, expressive unit that handles both linear and non-linear transformations natively.

\section{Background and Motivation}
In standard feedforward neural networks, a neuron performs a two-step process: it computes a weighted sum of inputs followed by the application of a non-linear activation function \cite{Szandała2021}. Mathematically, this is often written as shown in Equation~\ref{eq:traditional_neuron}.
\begin{equation}
y = \phi\left(\sum_{i=1}^{n} w_i x_i + b\right)
\label{eq:traditional_neuron}
\end{equation}
where $w_i$ are the trainable weights, $b$ is the bias term, and $\phi(\cdot)$ is a non-linear activation function such as ReLU, Tanh, Swish, Mish, or APTx activation function.

While this modular design offers flexibility, it also imposes structural redundancy and increased memory overhead. The separation between linear and non-linear components requires additional layers and parameters, making it less efficient, especially in memory-constrained or latency-critical environments.

Moreover, traditional neurons rely on fixed activation functions that remain the same across the network or layer. This rigid formulation limits the network's ability to adapt activation behavior dynamically based on the input distribution or training dynamics.

\medskip
  
The APTx activation function \cite{kumar2022aptx,aptxactivation2022code}, due to its parametric nature, already offers adaptive behavior and a rich expressiveness to simulate or interpolate between several standard non-linearities. It introduces trainable parameters $\alpha$, $\beta$, and $\gamma$ into the activation itself, allowing it to learn optimal non-linear transformations during training. This results in improved flexibility and performance across tasks and architectures.

Extending this idea, we hypothesized that merging the activation and computation stages into a single unit, while preserving full trainability, can lead to more compact and powerful architectures. Instead of separating a weighted summation and a subsequent activation, we propose a unified formulation that naturally incorporates both. This approach eliminates the need for explicit activation layers, reduces parameter duplication, and potentially enhances representational efficiency.

This line of thinking led to the design of the APTx Neuron, described in the next section.

\section{APTx Neuron}

The APTx Neuron is a novel computational unit that unifies linear transformation and non-linear activation into a single, expressive formulation. Inspired by the parametric APTx activation function, this neuron architecture removes the strict separation between computation and activation, allowing both to be learned as a cohesive entity. It is designed to enhance representational flexibility while reducing architectural redundancy. A key theoretical property of the APTx Neuron is that it naturally retains the universal approximation capability~\cite{HORNIK1989359}.

\subsection{Mathematical Formulation}

Traditionally, a neuron computes the output as shown in Equation \ref{eq:traditional_neuron_layer}.
\begin{equation}
y = \phi\left( \sum_{i=1}^{n} w_i x_i + b \right)
\label{eq:traditional_neuron_layer}
\end{equation}
where $x_i$ are the inputs, $w_i$ are the weights, $b$ is the bias, and $\phi$ is an activation function such as ReLU, Swish, or Mish.

The APTx Neuron merges these components into a unified trainable expression, as shown in Equation \ref{eq:aptx_neuron}.
\begin{equation}
y = \sum_{i=1}^{n} \left[ (\alpha_i + \tanh(\beta_i x_i)) \cdot \gamma_i x_i \right] + \delta
\label{eq:aptx_neuron}
\end{equation}
where:
\begin{itemize}
    \item $x_i$ is the $i$-th input feature,
    \item $\alpha_i$, $\beta_i$, and $\gamma_i$ are trainable parameters for each input,
    \item $\delta$ is a trainable scalar bias.
\end{itemize}

This equation allows the neuron to modulate each input through a learned, per-dimension non-linearity and scaling operation. The term $(\alpha_i + \tanh(\beta_i x_i))$ introduces adaptive gating, and $\gamma_i x_i$ provides multiplicative control.

\subsection{Relationship to Traditional Neurons and Activations}

A unique property of the APTx Neuron is its ability to represent multiple computational regimes depending on parameter values:
\begin{itemize}
   \item \textbf{Linear Neuron Equivalence:} If $\beta_i = 0$, then $\tanh(\beta_i x_i) = 0$, and the equation reduces to:
    \begin{equation}
    y = \sum_{i=1}^{n} (\alpha_i \cdot \gamma_i x_i) + \delta
    \end{equation}
    If we further assume either $\gamma_i = 1$ or $\alpha_i = 1$, then it becomes:
    \begin{equation}\label{eqn:uat_equivalence}
    y = \sum_{i=1}^{n} (\alpha_i \cdot x_i) + \delta \quad \text{or} \quad y = \sum_{i=1}^{n} (\gamma_i \cdot x_i) + \delta
    \end{equation}
    The form presented in Equation~\ref{eqn:uat_equivalence} is structurally identical to a conventional linear neuron with learnable weights and bias. Consequently, the APTx Neuron can recover the computation of a conventional linear neuron as a special case and may serve as a drop-in replacement for a conventional linear neuron within a neural-network architecture that possesses the universal approximation property~\cite{HORNIK1989359}.
    
    \item \textbf{Pure Activation Behavior:} When $\delta = 0$ and input is passed as a fixed vector (e.g., $x = [x_1, x_2, ..., x_n]$), the APTx Neuron computes a non-linear transformation akin to a composite APTx activation function as shown in Equation~\ref{eqn:composite_aptx_activation}.
    \begin{equation}\label{eqn:composite_aptx_activation}
    y = \sum_{i=1}^{n} \left[ (\alpha_i + \tanh(\beta_i x_i)) \cdot \gamma_i x_i \right]
    \end{equation}
    
    \item \textbf{Identity Function:} When $\alpha_i = 1/\gamma_i $, and $\beta_i = 0$, the neuron becomes:
    \begin{equation}
    y = \sum_{i=1}^{n} x_i + \delta
    \end{equation}
    again acting as a simple summing neuron with bias, matching standard behavior.
\end{itemize}
Because $\alpha_i$, $\beta_i$, $\gamma_i$, and $\delta$ are all trainable, the APTx Neuron can automatically adjust its role during training. It may act as a linear unit in some regions of the input space, and as a highly non-linear one in others.

\subsection{Design Benefits}

The unified formulation of APTx Neuron offers several practical and theoretical advantages:
\begin{itemize}
    \item \textbf{Expressive Adaptivity:} Each input dimension has its own dynamic non-linearity and transformation, enabling fine-grained learning.
    \item \textbf{Reduced Structural Complexity:} APTx Neurons eliminate the need for separate activation layers in hidden layers by integrating non-linearity within the neuron itself.
    \item \textbf{Enhanced Generalization:} The increased modeling freedom can help the APTx Neuron learn more compact representations without sacrificing performance.
    \item \textbf{Parameter Reusability:} APTx Neurons are capable of mimicking multiple types of traditional neurons, eliminating the need for hand-picking activation functions.
\end{itemize}

\subsection{Parameter Overhead and Efficiency}

Each APTx Neuron introduces $3n + 1$ trainable parameters for an input dimension $n$, compared to $n + 1$ in a standard neuron. However, due to their richer expressiveness, fewer APTx Neurons or layers may be required to achieve comparable or better performance. This often leads to a favorable trade-off between parameter count and model accuracy.

Although the original formulation uses parameters $\alpha_i$, $\beta_i$, and $\gamma_i$ for each input dimension $x_i$ within a neuron, along with a shared parameter $\delta$, several parameter sharing strategies can be adopted to reduce model complexity:
\begin{itemize}
    \item \textbf{Full sharing:} All parameters are shared and trainable across input dimensions, i.e., $\alpha_i = \alpha$, $\beta_i = \beta$, $\gamma_i = \gamma$, and a shared, trainable $\delta$.
    \item \textbf{Partial sharing:} For example, $\alpha_i = \alpha$ (shared and trainable), while $\beta_i$ and $\gamma_i$ remain input specific and trainable; $\delta$ remains shared and trainable.
    \item \textbf{Hybrid schemes:} Certain parameters may be fixed. For instance, setting $\alpha_i = 1$ (non-trainable), while keeping $\beta_i$, $\gamma_i$, and $\delta$ trainable. For a neuron with input dimension $n$, this reduces the total number of trainable parameters from $3n + 1$ to $2n + 1$.

\end{itemize}

These variants offer configurable trade-offs between parameter efficiency and expressive power, allowing flexible adaptation of the APTx Neuron to a variety of architectures and tasks.

Importantly, the formulation of the APTx Neuron preserves the universal approximation capability~\cite{HORNIK1989359} of neural networks. By embedding trainable, input-wise non-linearities directly within each neuron, the APTx Neuron goes beyond traditional designs where approximation power depends on stacking fixed activations atop linear transformations. Each input dimension in the APTx Neuron can independently learn its own transformation behavior, both linear and non-linear, enabling more efficient, compact, and expressive modeling, even in shallow architectures.

Additionally, as shown in Equation~\ref{eq:aptx_neuron}, the APTx Neuron integrates trainable, $\tanh$-based non-linear modulation directly into the neuron formulation. This unified design allows input transformation and non-linear behavior to be learned jointly within each APTx Neuron, without requiring a separate activation function in the hidden layers.

\medskip

In general, the APTx Neuron offers a powerful, flexible, and theoretically grounded alternative to conventional neuron designs. Its ability to bridge between pure activation behavior, linear transformation, and hybrid modes makes it a strong candidate for building more efficient and expressive neural networks.

\section{Architecture, Training, and Results}
To evaluate the effectiveness of the proposed APTx Neuron, we implemented a custom fully connected feedforward neural network using APTx Neurons in PyTorch~\cite{10.5555/3454287.3455008}. This section outlines the design, training pipeline, and performance metrics on the MNIST dataset~\cite{6296535}.

\subsection{Neural Network Design}

The APTx Neuron-based feedforward neural network replaces conventional linear and activation layers with custom layers composed of multiple APTx Neurons. Each APTx Neuron unifies computation and non-linearity into a single trainable expression. In this configuration, all parameters $\alpha_i$, $\beta_i$, $\gamma_i$, and $\delta$ were trainable. The architecture used in our experiments is as follows:

\begin{itemize}
    \item \textbf{Input:} Flattened MNIST image of size $28 \times 28 = 784$.
    \item \textbf{Layer 1:} APTx Layer with 128 neurons.
    \item \textbf{Layer 2:} APTx Layer with 64 neurons.
    \item \textbf{Layer 3:} APTx Layer with 32 neurons.
    \item \textbf{Output Layer:} Fully-connected linear layer with 10 output classes.
\end{itemize}

The total number of trainable parameters in the APTx Neuron-based feedforward neural network used in our experiments was \textbf{332,330}. While the APTx layers unify activation and computation, a softmax function is applied to the final output layer to produce class probabilities in the classification task in the MNIST dataset~\cite{6296535}. 

The source code for the APTx Neuron-based architecture and the MNIST~\cite{6296535} experiment, implemented in Python~\cite{python} using the PyTorch library~\cite{10.5555/3454287.3455008}, is available at our GitHub repository~\cite{aptxneuron2025code}.
In addition, a pip-installable Python package implementing the APTx Neuron is available on both GitHub~\cite{aptx_neuron_package_github} and PyPI~\cite{aptx_neuron_package_pip}.
It can be installed using the following command: \texttt{pip install aptx\_neuron}.

\subsection{Training Details}
\begin{itemize}
    \item \textbf{Dataset:} MNIST handwritten digit dataset.
    \item \textbf{Optimizer:} Adam with initial learning rate $4 \times 10^{-3}$.
    \item \textbf{Scheduler:} StepLR with step size of 5 epochs and decay rate of 0.25.
    \item \textbf{Loss Function:} CrossEntropyLoss.
    \item \textbf{Epochs:} 20.
    \item \textbf{Batch Size:} 64 for training, 1000 for testing.
\end{itemize}

All experiments were conducted on a single device (CPU or CUDA depending on availability), and training and test accuracy values were recorded for each epoch.
\subsection{Performance Metrics}

\begin{figure}
\includegraphics[width=\textwidth]{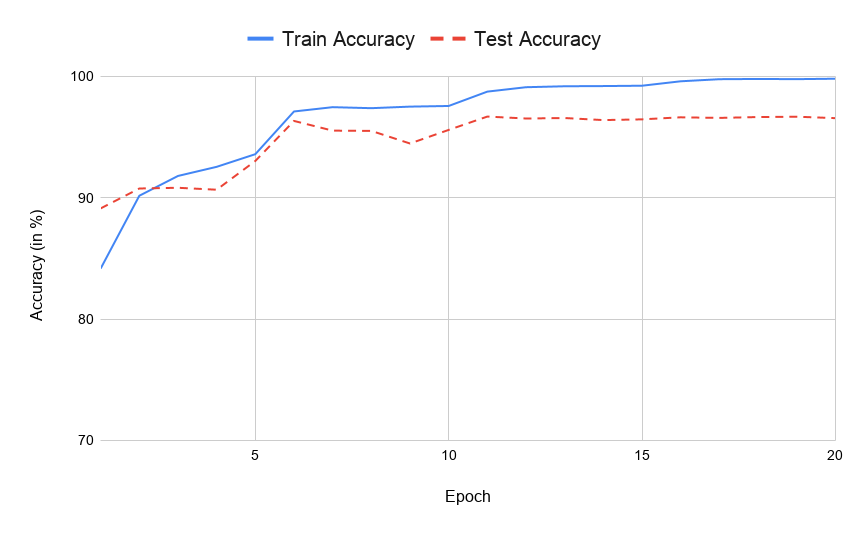}
\caption{Visual analysis of train and test accuracy values.} \label{fig1_accuracy}
\end{figure}

The APTx Neuron based model achieved a peak test accuracy of \textbf{96.69\%} at epoch 11. The training accuracy reached over \textbf{99.8\%} by epoch 20, indicating excellent capacity and convergence, as shown in Table \ref{tab:mnist_results}. Visual analysis of the training and test accuracy is shown in Figure~\ref{fig1_accuracy}.

\begin{table}[ht]
\centering
\caption{Performance of APTx Neuron-based feedforward neural network on MNIST.}
\label{tab:mnist_results}
\begin{tabular}{|c|c|c|}
\hline
\textbf{Epoch} & \textbf{Train Accuracy (\%)} & \textbf{Test Accuracy (\%)} \\
\hline
1 & 84.16 & 89.12 \\
2 & 90.16 & 90.76 \\
3 & 91.80 & 90.82 \\
4 & 92.55 & 90.66 \\
5 & 93.59 & 93.03 \\
6 & 97.11 & 96.33 \\
7 & 97.47 & 95.53 \\
8 & 97.38 & 95.51 \\
9 & 97.51 & 94.47 \\
10 & 97.56 & 95.59 \\
11 & 98.75 & \textbf{96.69} \\
12 & 99.11 & 96.53 \\
13 & 99.19 & 96.57 \\
14 & 99.21 & 96.40 \\
15 & 99.23 & 96.46 \\
16 & 99.60 & 96.63 \\
17 & 99.77 & 96.58 \\
18 & 99.79 & 96.65 \\
19 & 99.78 & 96.68 \\
20 & \textbf{99.81} & 96.56 \\
\hline
\end{tabular}
\end{table}

Although the APTx Neuron performs more operations per forward pass than a standard linear+activation unit, the model demonstrates rapid convergence, surpassing $96\%$ test accuracy by epoch 6.

\subsection{Insights}

\begin{itemize}
    \item APTx Neuron-based feedforward neural network converges quickly. It surpasses $96\%$ test accuracy within the first 6 epochs.
    \item The training accuracy reached $99.81\%$ by epoch 20, indicating strong fitting capacity on the MNIST training set.
    \item Despite a higher number of parameters per neuron, the overall architecture remains compact and efficient.
    \item The use of trainable non-linearities (APTx Neuron) enables superior representational power and dynamic learning behavior.
\end{itemize}

\section{Toward Integration in CNNs and Transformers}

The APTx Neuron, introduced in this work and defined in Equation~\ref{eq:aptx_neuron}, is a unified computational unit that integrates activation and transformation within a single trainable expression. Although its effectiveness has been demonstrated in fully connected feedforward networks, the formulation is not restricted to MLPs. As a general building block, the APTx Neuron and the APTx Activation Function (Equation~\ref{eq:aptx_activation}) can be incorporated into convolutional neural networks (CNNs) and Transformer architectures with minimal structural modification.

\subsection{Integration in Convolutional Neural Networks}

In convolutional neural networks (CNNs) \cite{726791}, activation functions such as ReLU, Swish, or Mish are commonly applied after convolutional filtering. The APTx framework supports two forms of integration: (i) replacing only the activation function, or (ii) forming a unified modulation operator on top of the convolutional responses, analogous to the modulation in Equation~\ref{eq:aptx_neuron}.

\subsubsection{Case 1: Activation-Only Replacement:}
A convolutional layer produces feature maps $z_k$ as defined in Equation~\ref{eqn:cnn_feature_map}.
\begin{equation}\label{eqn:cnn_feature_map}
z_k = w_k * x,\qquad z_k \in \mathbb{R}^{H \times W}
\end{equation}

where $C_{\text{out}}$ is the number of output channels of the convolutional layer and 
$H \times W$ denotes the spatial resolution (height and width) of each feature map.
The convolution kernels $w_k$ remain unchanged, and the APTx Activation Function from Equation~\ref{eq:aptx_activation} is applied element-wise as shown in Equation~\ref{eqn:cnn_aptx_activation}, where parameters $\alpha_k,\beta_k,\gamma_k$ are shared across spatial locations.
\begin{equation}\label{eqn:cnn_aptx_activation}
y_k(i,j) = \bigl(\alpha_k + \tanh(\beta_k z_k(i,j))\bigr)\,\gamma_k\, z_k(i,j)
\end{equation}
This produces the output $Y = [y_1,\dots,y_{C_{\text{out}}}] \in \mathbb{R}^{C_{\text{out}} \times H \times W}$.

\subsubsection{Case 2: Unified APTx Convolutional Unit:}
A more expressive integration applies an APTx-style modulation, similar to the structure in Equation~\ref{eq:aptx_neuron}, directly to the convolutional responses as expressed in Equation~\ref{eqn:cnn_aptx_neuron}.
\begin{equation}\label{eqn:cnn_aptx_neuron}
y_k(i,j) = \bigl(\alpha_k + \tanh(\beta_k z_k(i,j))\bigr)\,\gamma_k\, z_k(i,j) + \delta_k
\end{equation}
Depending on the learned parameters, the unit may behave like a dynamic activation or a channel-wise neuron-like modulator. The resulting output is $Y = [y_1,\dots,y_{C_{\text{out}}}] \in \mathbb{R}^{C_{\text{out}} \times H \times W}$.

\subsection{Integration in Transformer Architectures}

Transformers \cite{10.5555/3295222.3295349} rely on multi-head self-attention and a position-wise feed-forward network (FFN). Both components can benefit from APTx-based units.

\subsubsection{Replacing Q, K, and V Projections:}
The projections $Q$, $K$, and $V$ as shown in Equation~\ref{eqn:transformer_projections} may be replaced by \emph{APTxLayer} blocks constructed from APTx Neurons (Equation~\ref{eq:aptx_neuron}) as shown in Equation~\ref{eqn:transformer_projections_aptx_neuron}.
\begin{equation}\label{eqn:transformer_projections}
Q = W_Q x,\qquad K = W_K x,\qquad V = W_V x
\end{equation}

\begin{equation}\label{eqn:transformer_projections_aptx_neuron}
Q = \mathrm{APTxLayer}(x),\qquad 
K = \mathrm{APTxLayer}(x),\qquad 
V = \mathrm{APTxLayer}(x)
\end{equation}

\subsubsection*{Replacing the Feed-Forward Layer with an APTx Neuron Layer:}
The standard FFN is defined in Equation~\ref{eqn:ffn}.
\begin{equation}\label{eqn:ffn}
\mathrm{FFN}(x) = W_2\,\phi(W_1 x + b_1) + b_2
\end{equation}

Replacing $\phi(\cdot)$ with the APTx activation function as $(\alpha + \tanh(\beta u))\,\gamma u$, and letting $u = W_1 x + b_1$, we obtain Equation~\ref{eqn:ffn_aptx_activation}.
\begin{equation}\label{eqn:ffn_aptx_activation}
\mathrm{FFN}_{\mathrm{APTxActivation}}(x)
= W_2 \left( (\alpha + \tanh(\beta (W_1 x + b_1)))\,\gamma (W_1 x + b_1) \right) + b_2
\end{equation}

A full FFN block can also be constructed using two stacked APTxLayers,
each composed of APTx Neurons, as shown in Equation~\ref{eqn:ffn_aptx_neuron}. Since each APTx Neuron (Equation~\ref{eq:aptx_neuron}) already combines linear transformation, non-linearity, and bias, two such layers are sufficient to mirror the role of the standard two-layer FFN.

\begin{equation}\label{eqn:ffn_aptx_neuron}
\mathrm{FFN}_{\mathrm{APTxNeuron}}(x)
  = \mathrm{APTxLayer}_2\!\bigl(\mathrm{APTxLayer}_1(x)\bigr)
\end{equation}

\medskip
These integration pathways illustrate that APTx-based components form flexible and expressive building blocks for CNNs, Transformers, and hybrid neural architectures, extending the applicability of the unified APTx Neuron beyond multilayer perceptrons.

\section{Conclusion}
This work introduced the APTx Neuron, a unified, fully trainable neural unit that integrates linear transformation and non-linear activation into a single expression, extending the APTx activation function. By learning per-input parameters $\alpha_i$, $\beta_i$, and $\gamma_i$ for each input $x_i$, and a shared bias term $\delta$ within a neuron, the APTx Neuron removes the need for separate activation layers and enables fine-grained input transformation. The APTx Neuron generalizes traditional neurons and activations, offering greater representational power. Our experiments show that a fully connected APTx Neuron-based feedforward neural network achieves $96.69\%$ test accuracy on the MNIST dataset within 11 epochs using approximately 332K trainable parameters, demonstrating rapid convergence and high optimization efficiency. This design lays the groundwork for extending APTx Neurons to CNNs and transformers, paving the way for more compact and adaptive deep learning architectures.

\begin{credits}

\subsubsection{Competing Interest}
The authors have no competing interests to declare that are relevant to the content of this article.

\subsubsection*{Funding Information}
No institutional funding was received for this research.

\subsubsection*{Author Contribution}
All authors have contributed equally to the conception, implementation, analysis, and writing of the manuscript.

\subsubsection*{Data Availability Statement}
The source code and experimental results for this study are available at: \url{https://github.com/mr-ravin/APTxNeuron}. In addition, a pip-installable Python package implementing the APTx Neuron is available on both GitHub and PyPI:  
\begin{itemize}
    \item GitHub: \url{https://github.com/mr-ravin/aptx_neuron}
    \item PyPI: \url{https://pypi.org/project/aptx-neuron}
    \item Preprint on arXiv: \url{https://doi.org/10.48550/arXiv.2507.14270}
    \item Preprint on ssrn: \url{https://dx.doi.org/10.2139/ssrn.5364841}
\end{itemize}
The package can be installed using the following command:  
\texttt{pip install aptx-neuron}

\subsubsection*{Research Involving Human and/or Animals}
Not Applicable.

\subsubsection*{Informed Consent}
Not Applicable.

\end{credits}

%
%
%
\bibliographystyle{splncs04}
\bibliography{main}
\end{document}